\documentclass[runningheads]{llncs}
\usepackage{url}
\usepackage[normalem]{ulem}
\usepackage[T1]{fontenc}
\usepackage[nocompress]{cite}
\usepackage{amsfonts}
\usepackage{multicol}
\usepackage{multirow}
\usepackage{booktabs}
\usepackage{array}
\usepackage{arydshln}
\usepackage{xcolor}
\usepackage{colortbl}
\usepackage{algorithmicx}
\usepackage{subcaption}
\usepackage{algorithm}
\usepackage[noend]{algpseudocode}
\usepackage{amsmath}
\usepackage{amssymb}
\usepackage{tcolorbox}

\usepackage{graphicx,verbatim}
\usepackage[misc]{ifsym}

\begin{document}
\newcolumntype{P}[1]{>{\centering\arraybackslash}p{#1}}
\titlerunning{Continual Model Merging for WSI Survival}
\title{Benchmarking Data-Free Continual Model Merging for Whole-Slide Image Survival Analysis}
%

\author{Vu Minh Tran$^{1,2}$, Doanh C. Bui$^{1,2}$, Maï K. Nguyen$^{3}$, Khang Nguyen$^{1,2}$}
\authorrunning{Tran et al.}
\institute{$^1$University of Information Technology \\ $^2$Viet Nam National University Ho Chi Minh City \\ $^3$ETIS (UMR 8051), CY Cergy Paris University, ENSEA, CNRS, France
}

    
\maketitle    

\pagestyle{empty}

\begin{abstract}
Continual survival analysis on whole-slide images (WSIs) requires models to integrate knowledge from sequentially arriving cancer cohorts while preserving previously learned information. Model merging offers a data-free alternative by combining independently fine-tuned task experts without retaining previous training data. However, the relative behavior of different merging strategies for WSI-based survival prediction remains insufficiently understood. In this work, we establish an end-to-end benchmark for data-free continual model merging, covering task-specific fine-tuning, sequential merging, and inference under both TASK-IL and CLASS-IL settings. We systematically evaluate six representative merging strategies across four TCGA cancer cohorts and three task orders using C-index, IPCW C-index, Forgetting, and Backward Transfer under a controlled protocol. Results show that no merging strategy consistently dominates across all evaluation criteria, while censoring-aware evaluation can substantially alter the relative ranking of the compared methods. This benchmark provides a unified foundation for developing and evaluating more effective continual model merging methods for WSI-based survival analysis.


\end{abstract}

\section{Introduction}
Survival prediction from whole-slide images (WSIs) is an important task in computational pathology, supporting prognosis estimation and risk stratification for cancer patients~\cite{yao2020wsi,chen2021mcat}. However, existing WSI survival models are typically developed independently for individual cancer cohorts, requiring a new task-specific model whenever a new cohort is introduced~\cite{yao2020wsi,chen2021mcat}. This paradigm becomes increasingly costly in a continual setting, particularly for gigapixel-scale WSIs that demand substantial computational and storage resources~\cite{lu2021clam,chen2021mcat}. Continual learning provides a natural alternative by enabling a model to acquire new knowledge sequentially while retaining previously learned information~\cite{delange2022continual}.

Existing continual learning approaches commonly rely on regularization or rehearsal, with the latter requiring access to stored samples from previous tasks~\cite{delange2022continual}. Model merging offers a data-free alternative by independently fine-tuning task-specific experts from a shared pretrained model and combining their parameters without revisiting the original training data~\cite{wortsman2022modelsoups,ilharco2023taskarithmetic,yadav2023ties}. Recent work has demonstrated the feasibility of merging-based continual learning for WSI classification and survival prediction~\cite{bui2026mergeslide,mergesurv}. However, different merging strategies employ substantially different mechanisms for combining task-specific updates, and their relative behavior in continual WSI survival analysis remains insufficiently understood. This is particularly relevant for survival prediction, where right-censored outcomes require both conventional and censoring-aware evaluation~\cite{harrell1982yield,uno2011cstatistics}, while inference may be performed under either known or unknown task identity. A systematic benchmark under a common training, merging, and inference protocol is therefore needed to enable a fair comparison of these strategies.

To address this gap, we establish an end-to-end benchmark for data-free continual model merging in WSI-based survival analysis. The benchmark standardizes the pipeline from independent task-specific fine-tuning with a shared pretrained TITAN aggregator~\cite{ding2025titan}, through sequential model merging, to survival inference under both TASK-IL and CLASS-IL. The same independently trained task experts are reused across all task orders, allowing the effect of the merging strategy to be evaluated independently of task-specific training.

The main contributions of this work are summarized as follows:
\begin{itemize}
    \item We establish an end-to-end benchmark for data-free continual model merging in WSI-based survival analysis, covering task-specific fine-tuning, sequential merging, and inference under both TASK-IL and CLASS-IL.
    \item We systematically evaluate six representative merging strategies across four TCGA cohorts and three task orders using both conventional and censoring-aware survival metrics, together with continual learning measures, under a controlled protocol.
\end{itemize}

\section{Related Work}
\subsubsection{Pathology Foundation Models.}
Pretraining on large-scale histopathology data enables pathology foundation models to learn generalizable representations that can be transferred to diverse downstream tasks~\cite{lu2024conch,ding2025titan}. Recent models have extended this paradigm from patch-level visual representation learning to vision-language and slide-level modeling~\cite{lu2024conch,ding2025titan}. For example, CONCH provides transferable image representations for multiple computational pathology tasks~\cite{lu2024conch}, while TITAN further supports multimodal whole-slide representation learning and slide-level analysis ~\cite{ding2025titan}. These models provide a strong initialization for WSI survival prediction and continual adaptation, reducing the need to train task-specific feature extractors from scratch.

\subsubsection{Model Merging.}
Model merging aims to combine multiple independently fine-tuned models into a single model without requiring access to their original training data ~\cite{wortsman2022modelsoups,ilharco2023taskarithmetic,yadav2023ties}. This line of work has progressed from simple parameter averaging, as explored in Model Soups~\cite{wortsman2022modelsoups}, to task-vector-based approaches such as Task Arithmetic ~\cite{ilharco2023taskarithmetic}, and more recent conflict-aware methods including TIES-Merging ~\cite{yadav2023ties}. These approaches have demonstrated that task-specific knowledge can be integrated directly in parameter space, providing an efficient alternative to joint training or retraining. However, most conventional model merging methods are designed for a fixed set of models rather than a continually expanding task sequence~\cite{wortsman2022modelsoups,ilharco2023taskarithmetic,yadav2023ties}.

\subsubsection{Continual Model Merging.}
Continual model merging extends model merging to sequential settings, where task-specific models arrive over time and are progressively integrated without revisiting previous task data ~\cite{tang2025opcm,yang2025dop,qiu2026nufilt}. Recent studies have introduced dedicated approaches for this setting, including OPCM~\cite{tang2025opcm}, DOP~\cite{yang2025dop}, and NUFILT~\cite{qiu2026nufilt}, with the common goal of incorporating new task knowledge while preserving information already accumulated in the merged model. Although continual model merging has received increasing attention, existing studies have been evaluated predominantly on classification tasks~\cite{tang2025opcm,yang2025dop,qiu2026nufilt}. Its behavior in WSI-based survival analysis, where censored outcomes and survival-specific evaluation must be considered, remains insufficiently studied.

\section{Methodology}
\subsection{Problem formulation}
We consider a sequence of $N$ WSI-based survival tasks $\mathcal{D}=\{\mathcal{D}_t\}_{t=1}^{N}$, where each task corresponds to a distinct cancer cohort. For task $t$, $\mathcal{D}_t=\{(\mathcal{W}_i^t,Y_i^t)\}_{i=1}^{n_t}$, where $\mathcal{W}_i^t$ denotes WSI sample $i$ and $Y_i^t=(T_i^t,\delta_i^t)$ denotes its associated patient-level survival outcome. Here, $T_i^t$ is the observed survival time and $\delta_i^t\in\{0,1\}$ is the event indicator, with $\delta_i^t=1$ indicating an observed event and $\delta_i^t=0$ indicating right censoring. 
Let $f(\cdot;\theta_0)$ denote the pretrained TITAN WSI aggregator~\cite{ding2025titan}. Independent fine-tuning on task $t$ produces a task-specific aggregator $f(\cdot;\theta_t)$ and a Cox survival head $h(\cdot;\phi_t)$. Given a task arrival order $\pi=(\pi_1,\ldots,\pi_N)$, a continual model merging method $\mathcal{M}$ produces the merged aggregator after step $s$ as:
\begin{equation} 
\widetilde{\theta}_s = \mathcal{M}\left( \theta_0, \theta_{\pi_1},\ldots,\theta_{\pi_s} \right), \qquad s=1,\ldots,N. 
\end{equation} 
Only the aggregator parameters are merged, while the task-specific survival heads $\{\phi_t\}$ are retained for inference. During model merging, no training data from previously observed tasks are accessed.

\subsection{Overview of the Benchmark Pipeline}

\begin{figure}[!t]
    \centering
    \includegraphics[width=\linewidth]{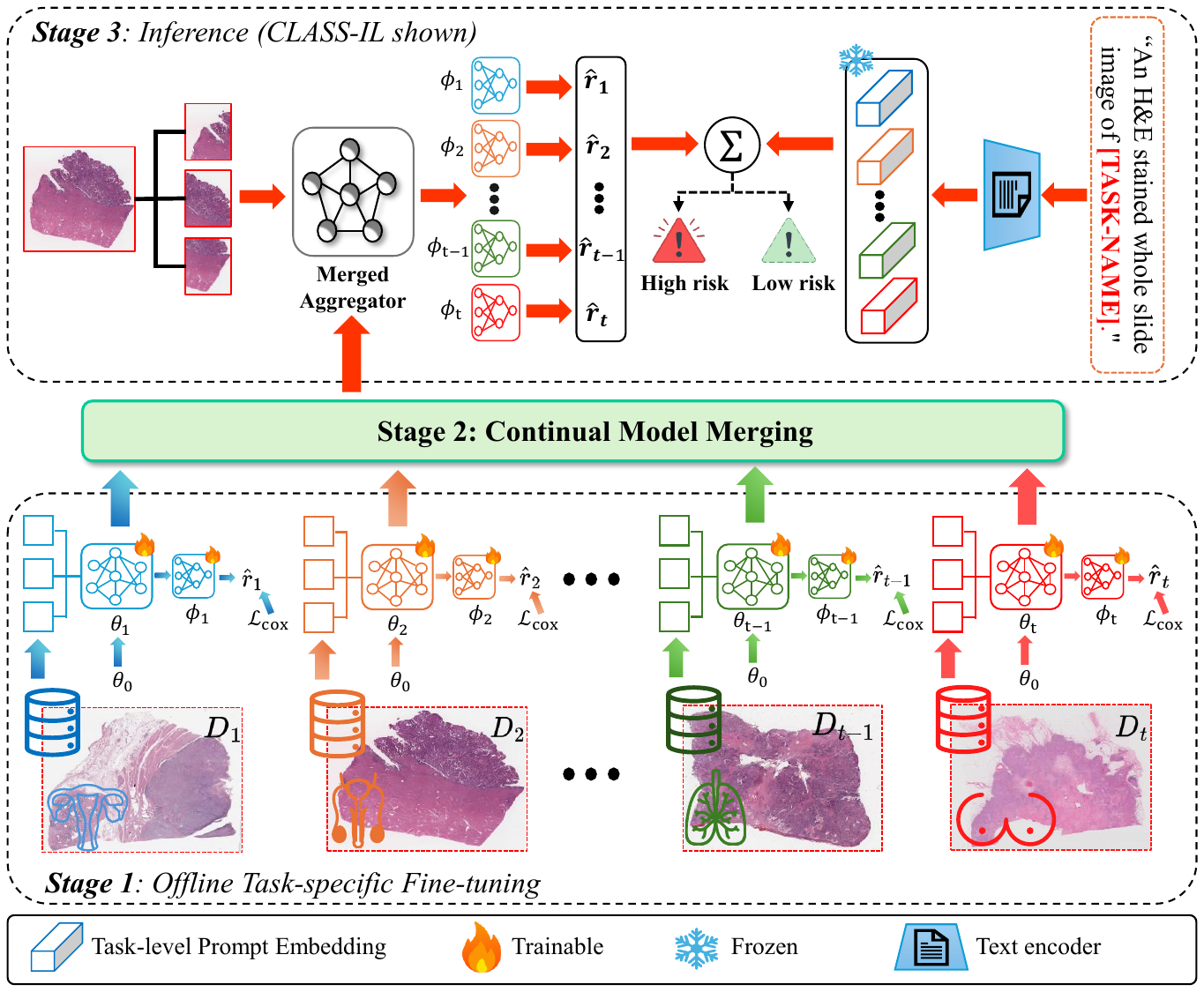}
    \caption{Overview of the proposed end-to-end benchmark pipeline for data-free continual model merging in WSI-based survival analysis.}
    \label{fig:pipeline}
\end{figure}

Our benchmark follows the three-stage pipeline illustrated in Fig.~\ref{fig:pipeline}: independent task-specific fine-tuning, sequential continual model merging, and survival inference under TASK-IL and CLASS-IL. Only the WSI aggregator is merged, while task-specific survival heads are retained for inference.

\subsection{Stage 1: Offline Task-specific Fine-tuning}
For each survival task $\mathcal{D}_t$, we independently fine-tune the WSI aggregator from the same pretrained TITAN parameters $\theta_0$. Each WSI $\mathcal{W}_i^t$ is represented by a bag of patch embeddings $X_i^t=\{\mathbf{x}_{i,k}^t\}_{k=1}^{K_i}$, which is processed by the task-specific aggregator and Cox survival head to predict $\hat r_i^t=h(f(X_i^t;\theta_t);\phi_t)$. The aggregator and survival head are jointly optimized using the Cox partial likelihood loss~\cite{cox1972}:
\begin{equation}
\mathcal{L}_{\mathrm{Cox}}^{t}
=
-\frac{1}{N_e^t}
\sum_{i:\delta_i^t=1}
\left[
\hat{r}_i^t
-
\log
\sum_{j:T_j^t\geq T_i^t}
\exp(\hat{r}_j^t)
\right],
\end{equation}
where $N_e^t=\sum_i\delta_i^t$ is the number of observed events. This yields the task-specific parameters $(\theta_t,\phi_t)$ for each survival task.

\subsection{Stage 2: Continual Model Merging Protocol}
In Stage 2, we benchmark six representative model merging methods: Weight Averaging (WA), Task Arithmetic (TA), TIES-Merging, Dual Orthogonal Projection (DOP), Null-space Filtering (NUFILT), and Orthogonal Projection-based Continual Merging (OPCM). Given a predefined task sequence, the task-specific TITAN aggregators obtained in Stage 1 are progressively integrated according to each merging strategy, while the corresponding survival heads are kept unchanged for inference.

\subsubsection{Weight Averaging (WA).}
Weight Averaging ~\cite{wortsman2022modelsoups} directly averages the parameters of all task-specific aggregators observed up to the current merging step. Given a task order $\pi=(\pi_1,\ldots,\pi_N)$, the merged aggregator after step $s$ is defined as
\begin{equation}
\widetilde{\theta}_s = \frac{1}{s}\sum_{j=1}^{s}\theta_{\pi_j},
\end{equation}
where $\theta_{\pi_j}$ denotes the fine-tuned aggregator parameters of task $\pi_j$. WA provides a simple parameter-space baseline without requiring additional optimization or task-specific hyperparameters.

\subsubsection{Task Arithmetic (TA).}
Task Arithmetic ~\cite{ilharco2023taskarithmetic} represents each fine-tuned aggregator as a task vector relative to the pretrained model. For task $t$, the task vector is defined as: $\tau_t = \theta_t - \theta_0,$ where $\theta_0$ denotes the pretrained TITAN aggregator parameters. After observing tasks up to step $s$, the merged aggregator is obtained by accumulating their task vectors:
\begin{equation}
\widetilde{\theta}_s = \theta_0 + \lambda \sum_{j=1}^{s}\tau_{\pi_j},
\end{equation}
where $\lambda$ controls the contribution of the accumulated task updates.

\subsubsection{TIES-Merging.}
TIES-Merging~\cite{yadav2023ties} reduces interference among task vectors through three operations:
trimming, sign resolution, and disjoint aggregation. At merging step $s$, each
observed task vector is first trimmed to retain its largest-magnitude entries:
\begin{equation}
\bar{\tau}_{\pi_j} = \operatorname{Trim}_{\rho}(\tau_{\pi_j}), \qquad j=1,\ldots,s,
\end{equation}

\noindent where $\rho$ denotes the retained proportion. A consensus sign vector is then
determined as:
\begin{equation}
\gamma_s = \operatorname{sign} \left(\sum_{j=1}^{s}\bar{\tau}_{\pi_j} \right).
\end{equation}

\noindent Only components consistent with the consensus sign are aggregated to obtain
$\tau_s^{\mathrm{TIES}}$. The merged aggregator is finally given by:
\begin{equation}
\widetilde{\theta}_s = \theta_0+\lambda\tau_s^{\mathrm{TIES}},
\end{equation}
where $\lambda$ controls the contribution of the merged task vector. In our
continual setting, TIES is reapplied to all task vectors observed up to each
merging step.

\subsubsection{Dual Orthogonal Projection (DOP).}
DOP~\cite{yang2025dop} performs sequential merging by jointly preserving the parameter-space knowledge of the previously merged aggregator and the incoming task aggregator. For a linear layer $\ell$, let $W_0^\ell$, $W_o^\ell=\tilde W_{s-1}^\ell$, and $W_n^\ell=W_{\pi_s}^\ell$ denote the pretrained, previously merged, and incoming task weights, respectively. Their updates with respect to the
pretrained model are
\begin{equation}
\Delta W_q^\ell=W_q^\ell-W_0^\ell,\qquad q\in\{o,n\}.
\end{equation}
DOP applies SVD to each task update, $\Delta W_q^\ell=U_q\Sigma_qV_q^\top$, and optimizes the merged weight by minimizing its projected discrepancy from both models:
\begin{equation}
\tilde W_s^\ell = \arg\min_W\sum_{q\in\{o,n\}}\omega_q\mathcal L_q(W),
\end{equation}
where
\begin{equation}
\mathcal L_q(W) = \left\| \Sigma_qU_q^\top(W-W_q^\ell) \right\|_F^2 + \left\|(W-W_q^\ell)V_q\Sigma_q\right\|_F^2.
\end{equation}
The objective weights $\omega_q$ are adaptively determined using multi-gradient descent to balance the two merging objectives. This procedure is applied to the linear weights of the TITAN aggregator, while the remaining
parameters are combined by simple averaging.

\subsubsection{Null-space Filtering (NUFILT).}
NUFILT~\cite{qiu2026nufilt} sequentially filters each incoming task update to reduce interference with the previously merged knowledge. For a selected linear layer $\ell$, let 
\begin{equation}
\Delta W_s^\ell = W_{\pi_s}^\ell-W_0^\ell, \qquad \Delta\widetilde{W}_{s-1}^\ell = \widetilde{W}_{s-1}^\ell-W_0^\ell
\end{equation}
denote the incoming and accumulated updates, respectively. SVD is applied to the accumulated update,
\begin{equation}
\Delta\widetilde{W}_{s-1}^\ell = U_{s-1}\Sigma_{s-1}V_{s-1}^{\top},
\end{equation}
and the retained right-singular basis $\bar{V}_{s-1}$ is used to construct the
null-space projector
\begin{equation}
P_s = I-\bar{V}_{s-1}\bar{V}_{s-1}^{\top}.
\end{equation}
NUFILT augments this projector with a learnable low-rank residual,
\begin{equation}
G_s=P_s+B_sA_s,
\end{equation}
and updates the merged weight as:
\begin{equation}
\widetilde{W}_s^\ell = \widetilde{W}_{s-1}^\ell + \Delta W_s^\ell G_s.
\end{equation}
The low-rank factors $A_s$ and $B_s$ are optimized to preserve the relevant
subspaces of both the incoming and previously merged updates.

\subsubsection{Orthogonal Projection-based Continual Merging (OPCM).}
Following MergeSurv~\cite{mergesurv}, we apply OPCM~\cite{tang2025opcm} to sequentially integrate the incoming task update into the previously merged aggregator. For a linear layer $\ell$, let
$\Delta\widetilde{W}_{s-1}^{\ell} =\widetilde{W}_{s-1}^{\ell}-W_0^{\ell}$
and
$\Delta W_s^{\ell}=W_{\pi_s}^{\ell}-W_0^{\ell}$ denote the accumulated and incoming updates, respectively. OPCM decomposes the accumulated update by SVD,
\begin{equation}
\Delta\widetilde{W}_{s-1}^{\ell}
=
U\Sigma V^{\top},
\end{equation}
projects the incoming update onto the resulting basis, $P=U^{\top}\Delta W_s^{\ell}V$, and suppresses its diagonal components:
\begin{equation}
\Delta W_{s,\mathrm{clean}}^{\ell} = U\left(P-\operatorname{Diag}(P)\right)V^{\top}.
\end{equation}
The filtered update is then integrated with the accumulated task vector using the adaptive norm balancing strategy of MergeSurv.

\subsection{Stage 3: Inference Protocols}
\subsubsection{TASK-IL.}
Under TASK-IL, the task identity is available at inference time. For a WSI represented by the patch-embedding bag $X_i^t$, the merged aggregator produces $\mathbf{z}_i^t=f(X_i^t;\tilde{\theta}_s)$, and the corresponding task-specific head predicts $\hat r_i^t=h(\mathbf{z}_i^t;\phi_t)$.
This setting directly evaluates the quality of the merged WSI representation when the correct survival head is known.

\subsubsection{CLASS-IL.}
For CLASS-IL, where task identity is unavailable, we adopt the Voting-Expert Aggregation (VEA) strategy from MergeSurv~\cite{mergesurv}.
Task-level prompt embeddings are generated using the TITAN text encoder ~\cite{ding2025titan}. Given the merged WSI representation $\mathbf{z}_i$, a routing weight for an
observed task $\pi_j$ is computed as:
\begin{equation}
p_{\pi_j}(\mathbf{z}_i) = \frac{
\exp\!\left(\cos(\mathbf{z}_i,\mathbf{e}_{\pi_j})/\gamma\right) }{\sum_{k=1}^{s}\exp\!\left(\cos(\mathbf{z}_i,\mathbf{e}_{\pi_k})/\gamma\right)},
\end{equation}
where $\mathbf{e}_{\pi_j}$ denotes the task-level text embedding and $\gamma$ is the routing temperature, set to $0.1$ in our experiments. The final survival risk is
\begin{equation}
\hat r_i =\sum_{j=1}^{s}p_{\pi_j}(\mathbf{z}_i)h(\mathbf{z}_i;\phi_{\pi_j}).
\end{equation}

\section{Evaluation Protocol}
\subsubsection{Datasets.}

\begin{table}[!t]
\centering
\caption{Statistics of the four TCGA cohorts.}
\label{tab:dataset_statistics}
\begin{tabular}{cccccc}
\toprule
Dataset
& Cancer Type
& WSIs
& Cens. WSIs
& Patients
& Cens. Patients \\
\midrule
TCGA-BLCA & Bladder      & 437  & 236 & 373 & 204 \\
TCGA-UCEC & Endometrial  & 539  & 460 & 480 & 405 \\
TCGA-LUAD & Lung         & 516  & 314 & 453 & 295 \\
TCGA-BRCA & Breast       & 1022 & 888 & 956 & 826 \\
\bottomrule
\end{tabular}
\end{table}

We evaluate WSI-based continual survival prediction on four cancer cohorts from The Cancer Genome Atlas (TCGA)~\cite{tcga2013}: Bladder Urothelial Carcinoma (BLCA), Uterine Corpus Endometrial Carcinoma (UCEC), Lung Adenocarcinoma (LUAD), and Breast Invasive Carcinoma (BRCA). Each cohort is treated as an individual continual learning task, with patient-level overall survival time and censoring information defining the prediction targets; cohort statistics are reported in Table~\ref{tab:dataset_statistics}. We use 5-fold patient-level cross-validation to prevent WSIs from the same patient from appearing across folds.

\subsubsection{Continual Task Orders.}
To assess the sensitivity of continual model merging methods to task arrival order, we evaluate all methods under three task sequences: \textcircled{\scriptsize 1} BLCA $\rightarrow$ UCEC $\rightarrow$ LUAD $\rightarrow$ BRCA; \textcircled{\scriptsize 2} BRCA $\rightarrow$ LUAD $\rightarrow$ UCEC $\rightarrow$ BLCA; \textcircled{\scriptsize 3} LUAD $\rightarrow$ BLCA $\rightarrow$ UCEC $\rightarrow$ BRCA. The same set of task-specific TITAN aggregators, independently fine-tuned in Stage 1, is reused across all three sequences, thereby isolating the effect of task order on merging performance.

\subsubsection{Metrics.}
We evaluate both survival prediction performance and continual learning behavior using C-index, IPCW C-index, Forgetting (FGT), and Backward Transfer (BWT).

\paragraph{\textbf{Concordance Index (C-index).}}
We use Harrell's concordance index~\cite{harrell1982yield} to evaluate the discriminative ability of the predicted survival risks. For WSI sample $i$ in task $t$, let $T_i^t$, $\delta_i^t$, and $\hat r_i^t$ denote the associated survival time, event indicator, and predicted risk score, respectively. A pair of WSI samples $(i,j)$ is considered comparable when $\delta_i^t=1$ and $T_i^t<T_j^t$. The C-index is defined as

\begin{equation}
C_t =
\frac{
\sum_{(i,j)\in\mathcal{P}_t}
\left[
\mathbb{I}(\hat{r}_i^t>\hat{r}_j^t)
+\frac{1}{2}\mathbb{I}(\hat{r}_i^t=\hat{r}_j^t)
\right]
}{
|\mathcal{P}_t|
},
\end{equation}
where $\mathcal{P}_t=\{(i,j)\mid \delta_i^t=1,\;T_i^t<T_j^t\}$ denotes the set of comparable pairs. A higher C-index indicates better agreement between the predicted risk ranking and the observed survival outcomes.

\paragraph{\textbf{IPCW C-index.}}
To account for right censoring, we additionally report the inverse probability of censoring weighted (IPCW) C-index~\cite{uno2011cstatistics}. Let $\hat{G}_t(\tau)$ denote the censoring survival function estimated from the training fold of task $t$ using the Kaplan--Meier estimator~\cite{kaplan1958}. For an observed event at time $T_i^t$, its contribution is weighted by $w_i^t=1/\hat{G}_t(T_i^t)^2$. The IPCW C-index is computed as
\begin{equation}
C_t^{\mathrm{IPCW}}
=
\frac{
\sum_{i,j}
w_i^t\,\delta_i^t\,
\mathbb{I}(T_i^t<T_j^t)
\left[
\mathbb{I}(\hat{r}_i^t>\hat{r}_j^t)
+\frac{1}{2}\mathbb{I}(\hat{r}_i^t=\hat{r}_j^t)
\right]
}{
\sum_{i,j}
w_i^t\,\delta_i^t\,
\mathbb{I}(T_i^t<T_j^t)
}.
\end{equation}
In our implementation, the truncation horizon $\tau$ is set to the maximum observed follow-up time in the training fold, and only observed events up to $\tau$ are considered. The censoring survival estimate is lower-bounded by $10^{-12}$ for numerical stability. A higher IPCW C-index indicates better censoring-adjusted concordance.

\paragraph{\textbf{Forgetting (FGT).}}
To quantify performance degradation on previously observed tasks during continual model merging, we follow the forgetting measure commonly used in continual learning~\cite{chaudhry2018rwalk} and define $A^{m}_{s,\pi_j}$ as the performance on task $\pi_j$ after merging up to step $s$, where $m\in\{C,C^{\mathrm{IPCW}}\}$. Forgetting is computed as
\begin{equation}
\mathrm{FGT}^{m}
=
\frac{1}{N-1}
\sum_{j=1}^{N-1}
\left(
\max_{s\in\{j,\ldots,N\}} A^{m}_{s,\pi_j}
-
A^{m}_{N,\pi_j}
\right).
\end{equation}
A lower FGT indicates better retention of previously acquired knowledge.

\paragraph{\textbf{Backward Transfer (BWT).}}
Backward transfer~\cite{lopezpaz2017gem} measures the effect of subsequent merging steps on previously observed tasks:
\begin{equation}
\mathrm{BWT}^{m}
=
\frac{1}{N-1}
\sum_{j=1}^{N-1}
\left(
A^{m}_{N,\pi_j}
-
A^{m}_{j,\pi_j}
\right).
\end{equation}
A positive BWT indicates improved performance on earlier tasks after incorporating later tasks, whereas a negative value indicates degradation.

\subsubsection{Implementation Details.}
\paragraph{\textbf{Training and Inference.}} WSIs are represented by pre-extracted CONCH v1.5 features~\cite{lu2024conch,ding2025titan} from $256\times256$ patches at $10\times$ magnification. Following the MIL setting~\cite{lu2021clam}, each WSI is represented as a bag of instances with $K=1000$ randomly sampled patches. Each task-specific TITAN aggregator is fine-tuned jointly with a Cox survival head with hidden dimension $512$ and dropout $0.25$. We train for $E=20$ epochs using AdamW with batch size $8$, learning rates of $10^{-5}$ and $10^{-4}$ for the aggregator and survival head, respectively, and weight decay $10^{-4}$.  Cox risk sets are constructed within each mini-batch, and batches without observed events contribute zero loss. All experiments are conducted on an NVIDIA A100 (80GB) GPU.

\paragraph{\textbf{Merging hyperparameters.}}
For TA, the task-vector scaling factor is set to $\lambda=0.3$. TIES uses $\lambda=0.3$, retains the top $20\%$ of task-vector entries by magnitude, and adopts sum-based disjoint merging. DOP is optimized for $200$ steps with a learning rate of $10^{-4}$, MGDA coefficient $\alpha=0.8$, EMA coefficient $\beta=0.999$, SVD threshold $\epsilon=0.99999$, and projection over both left- and right-singular subspaces (\texttt{uv}). NUFILT uses null-space filtering on \texttt{attn} and \texttt{fc1} linear layers, with a learning rate of $10^{-3}$, $50$ optimization steps, LoRA rank $64$, gradient rank $8$, and null-space rank $128$. For OPCM, we follow the configuration used in MergeSurv~\cite{mergesurv}.

\section{Results}

\begin{table}[!t]
\centering
\caption{Continual survival prediction results under the \textbf{TASK-IL} setting, averaged across three task orders \textcircled{\scriptsize 1} \textcircled{\scriptsize 2} \textcircled{\scriptsize 3}. For each order, mean and standard deviation are computed across five patient-level folds, with average performance evaluated at the final merging step. Best results are highlighted in \textbf{bold}, and the second-best results are \underline{underlined}.}
\label{tab:taskil_results}

\subcaption{\textbf{C-index}}
\resizebox{\textwidth}{!}{%
\begin{tabular}{lccc}
\toprule
\textbf{Method}
& \textbf{Average ($\uparrow$)}
& \textbf{FGT ($\downarrow$)}
& \textbf{BWT ($\uparrow$)} \\
\midrule

Weight Averaging (WA)
& 0.6525 ($\pm$0.0210)
& 0.0064 ($\pm$0.0064)
& \textbf{0.0050 ($\pm$0.0145)} \\

Task Arithmetic (TA)
& 0.6509 ($\pm$0.0195)
& 0.0114 ($\pm$0.0073)
& -0.0056 ($\pm$0.0112) \\

TIES-Merging (TIES)
& \uline{0.6533 ($\pm$0.0210)}
& \textbf{0.0052 ($\pm$0.0052)}
& \uline{0.0001 ($\pm$0.0085)} \\

DOP
& 0.6457 ($\pm$0.0232)
& 0.0084 ($\pm$0.0071)
& -0.0004 ($\pm$0.0127) \\

NUFILT
& \textbf{0.6543 ($\pm$0.0193)}
& \uline{0.0060 ($\pm$0.0045)}
& -0.0016 ($\pm$0.0068) \\

OPCM (MergeSurv)
& 0.6438 ($\pm$0.0227)
& 0.0155 ($\pm$0.0106)
& -0.0029 ($\pm$0.0179) \\

\bottomrule
\end{tabular}
}

\vspace{-0.2cm}

\subcaption{\textbf{IPCW C-index}}
\resizebox{\textwidth}{!}{%
\begin{tabular}{lccc}
\toprule
\textbf{Method}
& \textbf{Average ($\uparrow$)}
& \textbf{FGT ($\downarrow$)}
& \textbf{BWT ($\uparrow$)} \\
\midrule

Weight Averaging (WA)
& \uline{0.6132 ($\pm$0.0439)}
& 0.0121 ($\pm$0.0130)
& \uline{0.0084 ($\pm$0.0349)} \\

Task Arithmetic (TA)
& \textbf{0.6147 ($\pm$0.0643)}
& 0.0187 ($\pm$0.0195)
& -0.0060 ($\pm$0.0276) \\

TIES-Merging (TIES)
& 0.6130 ($\pm$0.0430)
& \textbf{0.0052 ($\pm$0.0054)}
& 0.0005 ($\pm$0.0100) \\

DOP
& 0.6035 ($\pm$0.0591)
& 0.0141 ($\pm$0.0144)
& -0.0020 ($\pm$0.0145) \\

NUFILT
& 0.6078 ($\pm$0.0443)
& \uline{0.0089 ($\pm$0.0136)}
& 0.0003 ($\pm$0.0181) \\

OPCM (MergeSurv)
& 0.6125 ($\pm$0.0625)
& 0.0175 ($\pm$0.0113)
& \textbf{0.0174 ($\pm$0.0364)} \\

\bottomrule
\end{tabular}
}
\end{table}

\begin{table}[!t]
\centering
\caption{Continual survival prediction results under the \textbf{CLASS-IL} setting,
averaged across three task orders \textcircled{\scriptsize 1} \textcircled{\scriptsize 2} \textcircled{\scriptsize 3}. For each order, mean and standard deviation are computed across five patient-level folds, with average performance evaluated at the final merging step. Best results are highlighted in \textbf{bold}, and the second-best results are \underline{underlined}.}
\label{tab:classil_results}

\subcaption{\textbf{C-index}}
\resizebox{\textwidth}{!}{%
\begin{tabular}{lccc}
\toprule
\textbf{Method}
& \textbf{Average ($\uparrow$)}
& \textbf{FGT ($\downarrow$)}
& \textbf{BWT ($\uparrow$)} \\
\midrule

Weight Averaging (WA)
& 0.6539 ($\pm$0.0186)
& 0.0072 ($\pm$0.0056)
& \textbf{0.0054 ($\pm$0.0112)} \\

Task Arithmetic (TA)
& 0.6523 ($\pm$0.0200)
& 0.0111 ($\pm$0.0064)
& -0.0050 ($\pm$0.0104) \\

TIES-Merging (TIES)
& \uline{0.6550 ($\pm$0.0195)}
& \uline{0.0062 ($\pm$0.0051)}
& 0.0007 ($\pm$0.0074) \\

DOP
& 0.6487 ($\pm$0.0266)
& 0.0087 ($\pm$0.0071)
& \uline{0.0024 ($\pm$0.0130)} \\

NUFILT
& \textbf{0.6571 ($\pm$0.0209)}
& \textbf{0.0058 ($\pm$0.0045)}
& 0.0002 ($\pm$0.0057) \\

OPCM (MergeSurv)
& 0.6436 ($\pm$0.0262)
& 0.0163 ($\pm$0.0093)
& -0.0030 ($\pm$0.0171) \\

\bottomrule
\end{tabular}
}

\vspace{-0.2cm}

\subcaption{\textbf{IPCW C-index}}
\resizebox{\textwidth}{!}{%
\begin{tabular}{lccc}
\toprule
\textbf{Method}
& \textbf{Average ($\uparrow$)}
& \textbf{FGT ($\downarrow$)}
& \textbf{BWT ($\uparrow$)} \\
\midrule

Weight Averaging (WA)
& \uline{0.6138 ($\pm$0.0432)}
& 0.0126 ($\pm$0.0133)
& \uline{0.0082 ($\pm$0.0344)} \\

Task Arithmetic (TA)
& 0.6116 ($\pm$0.0625)
& 0.0195 ($\pm$0.0189)
& -0.0089 ($\pm$0.0249) \\

TIES-Merging (TIES)
& \textbf{0.6152 ($\pm$0.0409)}
& \textbf{0.0043 ($\pm$0.0038)}
& 0.0012 ($\pm$0.0089) \\

DOP
& 0.6009 ($\pm$0.0597)
& 0.0132 ($\pm$0.0132)
& -0.0036 ($\pm$0.0144) \\

NUFILT
& 0.6089 ($\pm$0.0464)
& \uline{0.0077 ($\pm$0.0132)}
& 0.0009 ($\pm$0.0154) \\

OPCM (MergeSurv)
& 0.6065 ($\pm$0.0657)
& 0.0191 ($\pm$0.0112)
& \textbf{0.0119 ($\pm$0.0337)} \\

\bottomrule
\end{tabular}
}
\end{table}

\subsection{Overall Benchmark Performance}

Tables~\ref{tab:taskil_results} and~\ref{tab:classil_results} summarize the results aggregated across the three task orders. Under TASK-IL, NUFILT achieves the highest average C-index of $0.6543$, while TIES yields the lowest forgetting $0.0052$ and WA obtains the highest BWT $0.0050$. For the IPCW C-index, TA achieves the best average performance $0.6147$, whereas TIES again shows the lowest forgetting $0.0052$. OPCM obtains the highest IPCW-based BWT of $0.0174$.

Under CLASS-IL, NUFILT provides the strongest conventional C-index performance, achieving the highest average score $0.6571$ and the lowest forgetting $0.0058$, while WA achieves the highest BWT $0.0054$. In contrast, TIES performs best under IPCW evaluation, with the highest average IPCW C-index $0.6152$ and the lowest forgetting $0.0043$. OPCM again yields the highest BWT $0.0119$. Overall, the results show that no single merging method consistently dominates across predictive performance, knowledge retention, and backward transfer.

\begin{table}[!b]
\centering
\caption{Effect of task order on survival prediction performance. Results are averaged across five patient-level cross-validation folds. Best results within each inference setting and task order are highlighted in \textbf{bold}, and the second-best results are \underline{underlined}.}
\label{tab:task_order_results}
\resizebox{\textwidth}{!}{%
\begin{tabular}{cccccccc}
\toprule

\multirow{2}{*}{\textbf{Setting}}
& \multirow{2}{*}{\textbf{Method}}
& \multicolumn{3}{c}{\textbf{C-index}}
& \multicolumn{3}{c}{\textbf{IPCW C-index}} \\

\cmidrule(lr){3-5}
\cmidrule(lr){6-8}

&
& \textcircled{\scriptsize 1}
& \textcircled{\scriptsize 2}
& \textcircled{\scriptsize 3}
& \textcircled{\scriptsize 1}
& \textcircled{\scriptsize 2}
& \textcircled{\scriptsize 3} \\
\midrule

\multirow{6}{*}{TASK-IL}
& WA
& 0.6524
& 0.6527
& 0.6523
& \underline{0.6133}
& \underline{0.6136}
& 0.6127 \\

& TA
& 0.6508
& 0.6510
& 0.6509
& \textbf{0.6146}
& \textbf{0.6145}
& \textbf{0.6151} \\

& TIES
& \textbf{0.6533}
& \underline{0.6531}
& \underline{0.6535}
& 0.6129
& 0.6131
& \underline{0.6130} \\

& DOP
& 0.6452
& 0.6445
& 0.6473
& 0.6022
& 0.6039
& 0.6043 \\

& NUFILT
& \underline{0.6529}
& \textbf{0.6544}
& \textbf{0.6556}
& 0.6103
& 0.6034
& 0.6098 \\

& OPCM
& 0.6437
& 0.6440
& 0.6438
& 0.6118
& 0.6131
& 0.6126 \\

\midrule

\multirow{6}{*}{CLASS-IL}
& WA
& 0.6539
& 0.6540
& 0.6538
& \underline{0.6139}
& \underline{0.6141}
& \underline{0.6135} \\

& TA
& 0.6522
& 0.6524
& 0.6522
& 0.6119
& 0.6112
& 0.6118 \\

& TIES
& \underline{0.6548}
& \underline{0.6552}
& \underline{0.6550}
& \textbf{0.6151}
& \textbf{0.6152}
& \textbf{0.6152} \\

& DOP
& 0.6481
& 0.6452
& 0.6528
& 0.5985
& 0.5987
& 0.6056 \\

& NUFILT
& \textbf{0.6565}
& \textbf{0.6565}
& \textbf{0.6583}
& 0.6119
& 0.6037
& 0.6112 \\

& OPCM
& 0.6436
& 0.6437
& 0.6435
& 0.6064
& 0.6071
& 0.6061 \\

\bottomrule
\end{tabular}
}
\end{table}

\subsection{Impact of Censoring-aware Evaluation}
The choice of survival evaluation criterion substantially changes the relative ranking of continual model merging methods. Under TASK-IL, NUFILT achieves the highest conventional C-index at $0.6543$, followed by TIES at $0.6533$. However, after accounting for censoring through IPCW, TA becomes the best-performing method with an IPCW C-index of $0.6147$, while NUFILT drops to $0.6078$. A similar reversal appears under CLASS-IL: NUFILT achieves the highest conventional C-index of $0.6571$, whereas TIES ranks first under IPCW with $0.6152$, and NUFILT reaches only $0.6089$. These results reveal a clear metric-dependent trade-off: NUFILT is consistently favored by conventional discrimination, whereas TA and TIES become more competitive under censoring-aware evaluation. Consequently, no merging strategy is universally best across survival evaluation criteria, and the relative advantage of each method depends strongly on how censoring is incorporated into performance assessment.

\subsection{Robustness to Task Order}
Table~\ref{tab:task_order_results} evaluates the sensitivity of each merging method to the order in which tasks are introduced. Overall, the methods exhibit relatively small performance variations across the three task sequences, with the maximum range remaining below $0.01$ for both C-index and IPCW C-index. Among them, TIES shows particularly consistent behavior, with IPCW C-index varying by only $0.0002$ under TASK-IL and $0.0001$ under CLASS-IL. In contrast, larger fluctuations are observed for DOP and NUFILT, reaching $0.0076$ in CLASS-IL C-index and $0.0082$ in CLASS-IL IPCW C-index, respectively. These results indicate relatively low task-order sensitivity across the
evaluated sequences. However, strong predictive performance does not necessarily imply order robustness: NUFILT achieves the best conventional C-index but exhibits larger variation under censoring-aware evaluation, whereas TIES maintains competitive performance with consistently lower variation across task sequences.

\subsection{Discussion}
Our benchmark shows that no merging strategy is uniformly superior across survival evaluation criteria, suggesting that different merging mechanisms preserve and reconcile task-specific information in different ways. NUFILT performs strongest under the conventional C-index, whereas TA and TIES become more competitive under IPCW evaluation. One possible explanation lies in how the methods combine task updates: WA, TA, and TIES directly aggregate task-specific parameters or task vectors, while projection-based methods such as DOP, NUFILT, and OPCM rely on an accumulated merged state or subspace to constrain incoming updates. These different mechanisms may lead to distinct trade-offs between preserving shared knowledge and retaining task-specific information, which in turn become more apparent under different survival evaluation criteria. Our evaluation is limited to a single TITAN backbone with fixed patch sampling and four TCGA cohorts, leaving generalization to cross-center domain shifts and imbalanced task sequences unexplored. Future work will investigate broader settings and merging strategies that better balance conventional and censoring-aware survival performance.

\section{Conclusion}
We present an end-to-end benchmark for data-free continual model merging in WSI-based survival analysis, covering the full pipeline from independent task-specific fine-tuning to sequential model merging and survival inference under both TASK-IL and CLASS-IL settings. Using six representative merging strategies across multiple cancer cohorts and task orders, the benchmark evaluates predictive performance, forgetting, backward transfer, and censoring-aware discrimination within a unified protocol. The results show that no single merging strategy consistently dominates across all evaluation criteria, revealing complementary trade-offs among different merging mechanisms. This benchmark provides a reproducible foundation for systematic evaluation and future development of continual model merging methods for computational pathology.


%
%
%
\bibliographystyle{splncs04}
\bibliography{refs}

@article{yao2020wsi,
  author  = {Yao, Jiawen and Zhu, Xinliang and Jonnagaddala, Jitendra and Hawkins, Nicholas and Huang, Junzhou},
  title   = {Whole Slide Images Based Cancer Survival Prediction Using Attention Guided Deep Multiple Instance Learning Networks},
  journal = {Medical Image Analysis},
  volume  = {65},
  pages   = {101789},
  year    = {2020},
  doi     = {10.1016/j.media.2020.101789}
}

@inproceedings{chen2021mcat,
  author    = {Chen, Richard J. and Lu, Ming Y. and Weng, Wei-Hung and Chen, Tiffany Y. and Williamson, Drew F. K. and Manz, Trevor and others},
  title     = {Multimodal Co-Attention Transformer for Survival Prediction in Gigapixel Whole Slide Images},
  booktitle = {Proceedings of the IEEE/CVF International Conference on Computer Vision (ICCV)},
  pages     = {4015--4025},
  year      = {2021},
  doi       = {10.1109/ICCV48922.2021.00398}
}

@article{lu2021clam,
  author  = {Lu, Ming Y. and Williamson, Drew F. K. and Chen, Tiffany Y. and Chen, Richard J. and Barbieri, Matteo and Mahmood, Faisal},
  title   = {Data-Efficient and Weakly Supervised Computational Pathology on Whole-Slide Images},
  journal = {Nature Biomedical Engineering},
  volume  = {5},
  number  = {6},
  pages   = {555--570},
  year    = {2021},
  doi     = {10.1038/s41551-020-00682-w}
}

@article{delange2022continual,
  author  = {De Lange, Matthias and Aljundi, Rahaf and Masana, Marc and Parisot, Sarah and Jia, Xu and Leonardis, Ale{\v{s}} and others},
  title   = {A Continual Learning Survey: Defying Forgetting in Classification Tasks},
  journal = {IEEE Transactions on Pattern Analysis and Machine Intelligence},
  volume  = {44},
  number  = {7},
  pages   = {3366--3385},
  year    = {2022},
  doi     = {10.1109/TPAMI.2021.3057446}
}

@article{lu2024conch,
  author  = {Lu, Ming Y. and Chen, Bowen and Williamson, Drew F. K. and Chen, Richard J. and Liang, Ivy and Ding, Tong and others},
  title   = {A Visual-Language Foundation Model for Computational Pathology},
  journal = {Nature Medicine},
  volume  = {30},
  number  = {3},
  pages   = {863--874},
  year    = {2024},
  doi     = {10.1038/s41591-024-02856-4}
}

@article{ding2025titan,
  author  = {Ding, Tong and Wagner, Sophia J. and Song, Andrew H. and Chen, Richard J. and Lu, Ming Y. and Zhang, Andrew and others},
  title   = {A Multimodal Whole-Slide Foundation Model for Pathology},
  journal = {Nature Medicine},
  volume  = {31},
  number  = {11},
  pages   = {3749--3761},
  year    = {2025},
  doi     = {10.1038/s41591-025-03982-3}
}

@inproceedings{wortsman2022modelsoups,
  author    = {Wortsman, Mitchell and Ilharco, Gabriel and Gadre, Samir Ya and Roelofs, Rebecca and Gontijo-Lopes, Raphael and Morcos, Ari S. and others},
  title     = {Model Soups: Averaging Weights of Multiple Fine-Tuned Models Improves Accuracy Without Increasing Inference Time},
  booktitle = {Proceedings of the 39th International Conference on Machine Learning},
  series    = {Proceedings of Machine Learning Research},
  volume    = {162},
  pages     = {23965--23998},
  publisher = {PMLR},
  year      = {2022},
  url       = {https://proceedings.mlr.press/v162/wortsman22a.html}
}

@inproceedings{ilharco2023taskarithmetic,
  author    = {Ilharco, Gabriel and Ribeiro, Marco Tulio and Wortsman, Mitchell and Gururangan, Suchin and Schmidt, Ludwig and Hajishirzi, Hannaneh and others},
  title     = {Editing Models with Task Arithmetic},
  booktitle = {International Conference on Learning Representations},
  year      = {2023},
  url       = {https://openreview.net/forum?id=6t0Kwf8-jrj}
}

@inproceedings{yadav2023ties,
  author    = {Yadav, Prateek and Tam, Derek and Choshen, Leshem and Raffel, Colin A. and Bansal, Mohit},
  title     = {{TIES-Merging}: Resolving Interference When Merging Models},
  booktitle = {Advances in Neural Information Processing Systems},
  volume    = {36},
  pages     = {7093--7115},
  year      = {2023},
  doi       = {10.52202/075280-0310}
}

@inproceedings{bui2026mergeslide,
  author    = {Bui, Doanh C. and Ngo, Ba Hung and Pham, Hoai Luan and Nguyen, Khang and Nguyen, Ma{\"i} K. and Nakashima, Yasuhiko},
  title     = {{MergeSlide}: Continual Model Merging and Task-to-Class Prompt-Aligned Inference for Lifelong Learning on Whole Slide Images},
  booktitle = {2026 IEEE/CVF Winter Conference on Applications of Computer Vision (WACV)},
  pages     = {4859--4868},
  year      = {2026},
  doi       = {10.1109/WACV61042.2026.00472}
}

@misc{mergesurv,
  author        = {Tran, Vu Minh and Bui, Doanh C. and Pham, Van-Hau and Nguyen, Khang},
  title         = {{MergeSurv}: Merging-Based Continual Learning for Survival Analysis on Whole-Slide Images},
  year          = {2026},
  eprint        = {2607.04747},
  archivePrefix = {arXiv},
  primaryClass  = {cs.CV},
  note          = {arXiv:2607.04747},
  url           = {https://arxiv.org/abs/2607.04747}
}

@inproceedings{tang2025opcm,
  author    = {Tang, Anke and Yang, Enneng and Shen, Li and Luo, Yong and Hu, Han and Zhang, Lefei and others},
  title     = {Merging on the Fly Without Retraining: A Sequential Approach to Scalable Continual Model Merging},
  booktitle = {Advances in Neural Information Processing Systems},
  volume    = {38},
  pages     = {91233--91270},
  year      = {2025},
  doi       = {10.52202/085713-2747}
}

@inproceedings{yang2025dop,
  author    = {Yang, Enneng and Tang, Anke and Shen, Li and Guo, Guibing and Wang, Xingwei and Cao, Xiaochun and others},
  title     = {Continual Model Merging without Data: Dual Projections for Balancing Stability and Plasticity},
  booktitle = {Advances in Neural Information Processing Systems},
  volume    = {38},
  pages     = {43951--43981},
  year      = {2025},
  doi       = {10.52202/085713-1311}
}

@inproceedings{qiu2026nufilt,
  author    = {Qiu, Zihuan and Wang, Lei and Cao, Yang and Zhang, Runtong and Su, Bing and Xu, Yi and others},
  title     = {Null-Space Filtering for Data-Free Continual Model Merging: Preserving Stability, Promoting Plasticity},
  booktitle = {International Conference on Learning Representations},
  year      = {2026},
  url       = {https://proceedings.iclr.cc/paper_files/paper/2026/hash/d0da30e312b75a3fffd9e9191f8bc1b0-Abstract-Conference.html}
}

@article{cox1972,
  author  = {Cox, D. R.},
  title   = {Regression Models and Life-Tables},
  journal = {Journal of the Royal Statistical Society: Series B (Methodological)},
  volume  = {34},
  number  = {2},
  pages   = {187--202},
  year    = {1972},
  doi     = {10.1111/j.2517-6161.1972.tb00899.x}
}

@article{tcga2013,
  author  = {{The Cancer Genome Atlas Research Network} and Weinstein, John N. and Collisson, Eric A. and Mills, Gordon B. and Shaw, Kenna R. Mills and Ozenberger, Brad A. and others},
  title   = {{The Cancer Genome Atlas} Pan-Cancer Analysis Project},
  journal = {Nature Genetics},
  volume  = {45},
  number  = {10},
  pages   = {1113--1120},
  year    = {2013},
  doi     = {10.1038/ng.2764}
}

@article{harrell1982yield,
  author  = {{Harrell Jr.}, Frank E. and Califf, Robert M. and Pryor, David B. and Lee, Kerry L. and Rosati, Robert A.},
  title   = {Evaluating the Yield of Medical Tests},
  journal = {JAMA},
  volume  = {247},
  number  = {18},
  pages   = {2543--2546},
  year    = {1982},
  doi     = {10.1001/jama.1982.03320430047030}
}

@article{kaplan1958,
  author  = {Kaplan, E. L. and Meier, Paul},
  title   = {Nonparametric Estimation from Incomplete Observations},
  journal = {Journal of the American Statistical Association},
  volume  = {53},
  number  = {282},
  pages   = {457--481},
  year    = {1958},
  doi     = {10.1080/01621459.1958.10501452}
}

@article{uno2011cstatistics,
  author  = {Uno, Hajime and Cai, Tianxi and Pencina, Michael J. and D'Agostino, Ralph B. and Wei, L. J.},
  title   = {On the {C}-Statistics for Evaluating Overall Adequacy of Risk Prediction Procedures with Censored Survival Data},
  journal = {Statistics in Medicine},
  volume  = {30},
  number  = {10},
  pages   = {1105--1117},
  year    = {2011},
  doi     = {10.1002/sim.4154}
}

@inproceedings{chaudhry2018rwalk,
  author    = {Chaudhry, Arslan and Dokania, Puneet K. and Ajanthan, Thalaiyasingam and Torr, Philip H. S.},
  title     = {Riemannian Walk for Incremental Learning: Understanding Forgetting and Intransigence},
  booktitle = {Proceedings of the European Conference on Computer Vision (ECCV)},
  pages     = {532--547},
  year      = {2018},
  doi       = {10.1007/978-3-030-01252-6_33}
}

@inproceedings{lopezpaz2017gem,
  author    = {Lopez-Paz, David and Ranzato, Marc'Aurelio},
  title     = {Gradient Episodic Memory for Continual Learning},
  booktitle = {Advances in Neural Information Processing Systems},
  volume    = {30},
  pages     = {6467--6476},
  year      = {2017},
  url       = {https://proceedings.neurips.cc/paper/2017/hash/f87522788a2be2d171666752f97ddebb-Abstract.html}
}
%
\end{document}